\documentclass{article}

\usepackage{amsmath,amsfonts,bm}

\def\eqref#1{equation~\ref{#1}}

\def\1{\bm{1}}

\DeclareMathAlphabet{\mathsfit}{\encodingdefault}{\sfdefault}{m}{sl}
\SetMathAlphabet{\mathsfit}{bold}{\encodingdefault}{\sfdefault}{bx}{n}

\DeclareMathOperator*{\argmin}{arg\,min}

 \usepackage[final]{nips_2018}

\usepackage[utf8]{inputenc} 
\usepackage[T1]{fontenc}    
\usepackage{hyperref}       
\usepackage{url}            
\usepackage{booktabs}       
\usepackage{amsfonts}       
\usepackage{nicefrac}       
\usepackage{microtype}      
\usepackage{graphicx}
\usepackage{amsmath}
\usepackage{subcaption,enumitem, float}

\newcommand{\cut}[1]{}
\newcommand{\mbf}[1]{\mathbf{#1}}

\title{Modeling Irregularly Sampled Clinical Time Series}
\author{
Satya Narayan Shukla, Benjamin M. Marlin \\
  College of Information and Computer Sciences\\
University of Massachusetts Amherst\\
  Amherst, MA 01003 \\
  \texttt{\{snshukla,marlin\}@cs.umass.edu} \\
}

\begin{document}

\maketitle

\begin{abstract}
  While the volume of electronic health records (EHR) data continues to grow,
it remains rare for hospital systems to capture dense physiological
data streams, even in the data-rich intensive care unit setting. 
Instead, typical EHR records
consist of sparse and irregularly observed multivariate time series,
which are well understood to present particularly challenging problems
for machine learning methods. In this paper, we present a new deep learning
architecture for addressing this problem based on the use of 
a semi-parametric interpolation network followed by the 
application of a prediction network. 
The interpolation network allows for information to be shared across multiple
dimensions during the interpolation stage, while any standard
deep learning model can be used for the prediction network. 
We investigate the performance
of this architecture on the problems of mortality
and length of stay prediction. 
\end{abstract}

\section{Introduction}

While the volume of electronic health records (EHR) data continues to grow,
it remains rare for hospital systems to capture dense physiological
data streams, even in the data-rich intensive care unit setting. 
Instead, it is common for the physiological time series data in electronic health records
to be both sparse and irregularly sampled. An irregularly sampled time series 
is a sequence of samples with irregular intervals between their observation times.
Irregularly sampled data are considered to be sparse when the intervals
between successive observations are often large. The physiological
data contained in EHRs represent multivariate time series consisting
of the time course of multiple vital signs. In the multivariate setting,
the additional issue of the lack of alignment in the observation times
across physiological variables is also very common.

It is well understood that such data cause significant issues for standard 
supervised machine learning models that typically assume 
fully observed, fixed-size feature representations \citep{yadav2018mining}.
However, over the last several years, there has been significant progress 
in developing specialized models and architectures that can
accommodate sparse and irregularly sampled time series
as input \citep{marlin-ihi2012, li2015classification, li2016scalable, lipton2016directly, che2016recurrent, futoma2017improved}. Of particular interest in the supervised learning setting are 
methods that perform end-to-end learning directly using 
multivariate sparse and irregularly sampled time series as input without
the need for a separate interpolation or imputation step. 

In this work, we present a new model architecture for
supervised learning with multivariate sparse and irregularly sampled data: 
Interpolation-Prediction Networks. 
The architecture is based on the use of several semi-parametric interpolation layers
organized into an interpolation network, followed by the 
application of a prediction network that can leverage any standard deep learning model.

The interpolation network allows for information contained in each input time series
to contribute to the interpolation of all other time series
in the model. The parameters of the interpolation and prediction networks
are learned end-to-end via a composite objective function consisting 
of supervised and unsupervised components. The interpolation network serves 
the same purpose as the multivariate Gaussian process used in 
the work of \citet{futoma2017improved}, but remove the restrictions 
associated with the need for a positive definite covariance matrix. 

Our approach also allows us to compute an explicit multi-timescale representation
of the input time series, which we use to isolate information
about transients (short duration events) from broader trends.  
Similar to the work of \citet{lipton2016directly} and  \citet{che2016recurrent},
our architecture also explicitly leverages a separate information
channel related to the pattern of observed values. However, our
representation uses a semi-parametric intensity function representation
of this information that is more closely related to the work of \citet{lasko2014efficient}
on modeling medical event point processes.


We evaluate the proposed architecture on two tasks using the MIMIC-III
data set \citep{johnson2016mimic}: mortality prediction and length of 
stay prediction. Our approach outperforms a variety of simple baseline
models as well as the basic and advanced GRU models introduced by 
\citet{che2016recurrent} on both tasks across several metrics. 

\section{Model Framework}

We let $\mathcal{D}=\{(\mbf{s}_i,y_i)|i=1,...,N\}$ represent a data set containing
$N$ data cases. An individual data case consists of
a single target value $y_n$ (discrete in the case of classification
and real-valued in the case of regression), as well as a $D$-dimensional, sparse and irregularly
sampled multivariate time series $\mbf{s}_n$. Different dimensions $d$ of the multivariate
time series can have observations at different times, as well as different total
numbers of observations $L_{dn}$. Thus, we represent time series $d$ for data case $n$ as a tuple
$\mbf{s}_{dn}=(\mbf{t}_{dn}, \mbf{x}_{dn})$  where $\mbf{t}_{dn}=[t_{1dn},...,t_{L_{dn}dn}]$
is the list of time points at which observations are defined and 
$\mbf{x}_{dn}=[x_{1dn},...,x_{L_{dn}dn}]$ is the corresponding list of observed values.

The overall model architecture consists of two main components: an interpolation
network and a prediction network. The interpolation network interpolates
the multivariate, sparse, and irregularly sampled input time series against 
a set of reference time points $\mbf{r}=[r_1,...,r_T]$. 
In this work, we propose a two-layer interpolation network with each layer performing a different type of interpolation.


The first interpolation layer in the interpolation network performs a semi-parametric 
univariate interpolation for each of the $D$ time series separately. The interpolation
is based on a radial basis function network. This
results in a set of values $\hat{x}^{1c}_{kdn}$ for each data case $n$, each input
time series $d$, and each reference time point $r_k$, and each interpolation channel 
$c$ as shown in Equation \ref{eq:interp1}.
\vspace{-1mm}
\begin{align}
 \label{eq:interp1}
{ \hat{x}^{1c}_{kdn} = f^{1c}_{\theta}(r_k,\mbf{s}_n) = \frac{ \sum_{j=1}^{L_{dn}} w_{dc}(r_k,t_{jdn}) \:x_{jdn} }{\sum_{j=1}^{L_{dn}}  w_{dc}(r_k,t_{jdn})} }  \:, &&
  w_{dc}(r_k,t_{jdn}) = \exp\left(-\alpha_{dc} ||r_k - t_{jdn}||^2\right)
\end{align}

The second interpolation layer merges information across all of the $D$
time series at each reference time point by taking into account
learned correlations $\rho_{dd'}$ across all time series. This
results in a set of values $\hat{x}^{2c}_{kdn}$ for each data case $n$, each input
time series $d$, each reference time point $r_k$, and each interpolation channel 
$c$ as shown in Equation \ref{eq:interp2}. Here, the terms $i_{kd}$ represent
the intensity of the observations on input dimension $d$ for data  case $n$. The more 
observations that occur near reference time point $r_k$, the larger the value
of $i_{kdn}$. This final interpolation layer thus models the 
the interpolant $\hat{x}^{2c}_{kdn}$ as a weighted linear combination of the
first layer interpolants, while focusing the combination on the time series
$d'$ for which data are actually available at nearby time points.
\vspace{-1mm}
\begin{align}
 \label{eq:interp2}
 \hat{x}^{2c}_{kdn} = f^{2c}_{\theta}(r_k,\mbf{s}_n) =\frac{\sum_{d'} \rho_{dd'} \: i_{kd'n}^c \: \hat{x}^{1c}_{kd'n}}{\sum_{d'} i_{kd'n}^c}  \: , \qquad
 i_{kdn}^c =  f^{3c}_{\theta}(r_k,\mbf{s}_n) = \sum_j w_{dc}(r_k, t_{jdn}) 
\end{align}

In the experiments presented in the next section, we use a total of three
interpolation outputs ($C=3$) corresponding to a smooth interpolant to capture
trends, a non-smooth interpolant to capture transients, and the 
intensity function to retain information about where observations occur.
The smooth interpolant corresponds to the first interpolant 
$\hat{\mbf{x}}^{21}_{dn} = [\hat{x}^{21}_{1dn},...,\hat{x}^{21}_{Tdn}]$. 
The collection of model parameters associated with this interpolant are the 
cross-dimension correlation coefficients $\rho_{dd'}$, and the 
RBF network bandwidths $\alpha_{d1}$ for all $d$ and $d'$. 

For the non-smooth interpolant, we start with a second interpolant
$\hat{\mbf{x}}^{12}_{dn} = [\hat{x}^{12}_{1dn},...,\hat{x}^{12}_{Tdn}]$.
The collection of model parameters associated with this 
interpolant are the RBF network parameters $\alpha_{d2}$ for all $d$. 
To accomplish the goal of having this component represent a less smooth interpolation
than  $\hat{\mbf{x}}^{21}_{dn}$, we enforce the relationship 
$\alpha_{d2} = \kappa \alpha_{d1}$ for all $d$ for a value of $\kappa$ greater than one.
This ensures that the temporal similarity decays faster for the component intended to
model transients. To further minimize any redundancy between 
$\hat{\mbf{x}}^{21}_{dn}$ and $\hat{\mbf{x}}^{12}_{dn}$, we subtract the 
smooth interpolant from the non-smooth interpolant leaving the non-smooth
residual: $\hat{\mbf{x}}'^{12}_{dn}=\hat{\mbf{x}}^{12}_{dn}-\hat{\mbf{x}}^{21}_{dn}$.

Finally, for the intensity function, we use 
$\mbf{i}_{dn}^1 = [i^{1}_{1dn},...,i^{1}_{Tdn}]$. This component 
shares its RBF network parameters with  $\hat{\mbf{x}}^{21}_{dn}$.
In the experiments, we study the end-to-end impact of each
of these interpolation outputs. 

The second component, the prediction network, takes the output of the interpolation
network as its input and produces a prediction $\hat{y}_n$ for the target variable.
The prediction network can consist of any standard supervised 
neural network architecture (fully-connected feedforward, convolutional, recurrent, etc).
Thus, the architecture is fully modular with respect to the use of different prediction
networks. Appendix \ref{diagram} shows the architecture of the proposed model.


%

To learn the model parameters, we use a 
composite objective function consisting of a supervised component
and an unsupervised component. This is due to the fact that the 
supervised component alone is insufficient to learn reasonable
parameters for the interpolation network given the amount
of available training data. The unsupervised component used 
corresponds to an autoencoder-like loss function. However,
the semi-parametric RBF interpolation layers have the ability to
exactly fit the input points by setting the RBF kernel parameters
to very large values.  To avoid this solution and force the interpolation
layers to learn to properly interpolate the input data, it is necessary
to hold out some observed data points $x_{jdn}$ during learning and then
to compute the reconstruction loss only for these data points. This is a 
well-known problem with high-capacity autoencoders, and past work 
has used similar strategies to avoid the problem of trivially 
memorizing the input data without learning useful structure.

To implement the autoencoder component of the loss, we introduce
a set of masking variables $m_{jdn}$ for each data point $(t_{jdn}, x_{jdn})$.
If $m_{jdn}=1$, then we remove the data point $(t_{jdn}, x_{jdn})$ as an input
to the interpolation network, and include the predicted value of this time point
when assessing the autoencoder loss. We use the shorthand notation $\mbf{m}_n \odot \mbf{s}_n$ to represent
the subset of values of $\mbf{s}_n$ that are masked out, and $(1-\mbf{m}_n) \odot \mbf{s}_n$
to represent the subset of values of $\mbf{s}_n$ that are not masked out. 
The value $\hat{x}_{jdn}$ that we predict for a masked input at time  
point $t_{jdn}$ is the value of the smooth interpolant at that time point, calculated
based on the un-masked input values: $\hat{x}_{jdn} = f^{21}(t_{jdn}, (1-\mbf{m}_n) \odot \mbf{s}_n)$.

Using these definitions, we can now specify the learning problem for
the proposed framework. We let $\ell_P$ be the loss for the prediction
network (we use cross-entropy loss for classification and squared 
error for regression). We let $\ell_I$ be the interpolation network autoencoder loss (we use standard squared error).
We also include $\ell_2$ regularizers for both the interpolation
and prediction networks parameters.

\vspace{-7mm}
\begin{align}
	\theta_*,\phi_* &=\argmin_{\theta,\phi} \sum_{n=1}^N \ell_P(y_n, g_{\phi}(f_{\theta}(\mbf{s}_n))
	+ \lambda_I \Vert \theta \Vert_2^2 + \lambda_P \Vert \phi \Vert_2^2 \\
	\nonumber &\;\;\;\;\;\;\;\;\;\;\;\;+ \delta \sum_{n=1}^N \sum_{d=1}^D \sum_{j=1}^{L_{dn}} m_{jdn} \ell_I(x_{jdn},f^{21}(t_{jdn}, (1-\mbf{m}_n) \odot \mbf{s}_n))
\end{align}
where $f$ and $ g$ are interpolation and prediction network respectively.

\section{Experiments and Results}
We test the model framework on publicly available MIMIC-III  data set \footnote{MIMIC-III dataset is publicly available at \url{https://mimic.physionet.org/}}, a multivariate time series dataset consisting of sparse and irregularly sampled physiological signals \citep{johnson2016mimic}. More details on data extraction and sparsity are available in appendix \ref{dataset}. We use mortality and length of stay prediction as example classification and regression tasks.

We compare the proposed model with a number of baseline approaches including off-the-shelf classification and regression models learned using basic features, as well as more recent approaches based on customized neural network models. For non-neural network baselines, we evaluate Logistic Regression \citep{hosmer2013applied}, Linear Regression \citep{hastie01statisticallearning}, Support Vector Machines (SVM) \citep{cortes1995}, Random Forest (RF) \citep{breiman_rf} and AdaBoost \citep{adaboost}. Standard instances of all of these models require fixed-size feature representations. We use the mean of the available values in each of the physiological time series for a given admission record as the feature set. 
In addition, we compare to several existing deep learning baselines built on GRUs using simple interpolation or imputation approaches: GRU-M (missing observations replaced with global mean), 
GRU-F (missing observations replaced with last observation), GRU-S (input concatenated with masking variable to identify missingness and time interval indicating how long the particular variable is missing) and GRU-D \cite{che2016recurrent} (similar to GRU-S except instead of just replacing the missing value with the last measurement, it is decayed over time towards the empirical mean).  Training and implementation details can be found in appendix \ref{implementation}. 

We report the results from the 5-fold cross validation in terms of the average area under the ROC curve (AUC score), average area under precision-recall curve (AUPRC score) and average cross-entropy loss for classification task and average median absolute error and average fraction of 
explained variation score (EV) for regression task. We also report the standard deviation over cross validation folds. We note that in highly skewed datasets, as is the case with MIMIC-III, AUPRC  \citep{auprc} can give better insight into classification performance compared to the  AUC score.

\begin{table}[t]
\caption{Performance on mortality and length of stay prediction  tasks on MIMIC-III. Loss: Cross-Entropy Loss, MedAE: Median Absolute Error (in days), EV: Explained variance}
\footnotesize
\begin{center}
\begin{tabular}{ l c c c c c} 
 \toprule
 {\bf Model}&  \multicolumn{3}{c}{\bf Classification} &  \multicolumn{2}{c}{\bf Regression}\\
 \midrule
 {} & {\bf AUC} & {\bf AUPRC} & {\bf Loss} & {\bf MedAE} & {\bf EV score}\\
  \midrule
 {Log/LinReg} & $0.772\pm 0.013$ & $0.303\pm0.018$  & $0.240\pm 0.003$  & $3.528\pm0.072$&	$0.043\pm0.012$\\ 
  {SVM} & $0.671\pm 0.005$ & $0.300\pm0.011$& $0.260\pm 0.002$&	$3.523\pm0.071$&	$0.042\pm0.011$\\ 
 
  AdaBoost & $0.829\pm0.007$& $0.345\pm0.007$	&$0.663\pm0.000$ &	$4.517\pm0.234$& 	$0.100\pm0.012$\\ 
 
  RF & $0.826\pm 0.008$ &	$0.356\pm0.010$&	$0.315\pm0.025$ &	$3.113\pm0.125$&	$0.117\pm0.035$  \\ 
 
  GRU-M & $0.831\pm0.007$ & 	$0.376\pm0.022$&	$0.220\pm0.004$ &	$3.140\pm0.196$&	$0.131\pm0.044$\\ 
 
 GRU-F & $0.821\pm0.007$ &  $0.360\pm0.013$&		 $0.224\pm0.003$&	$3.064\pm0.247$& 	$0.126\pm0.025$ \\
 
  GRU-S & $0.843\pm0.007$ & 	$0.376\pm0.014$ &		$0.218\pm0.005$  &	$2.900\pm0.129$&	$0.161\pm0.025$\\ 

 GRU-D & $0.835\pm0.013$ & 	$0.359\pm0.025$&	$0.225\pm0.009$ &  ${\bf 2.891\pm0.103}$&	$0.146\pm0.051$\\

 {\bf Proposed } &  ${\bf 0.853\pm 0.007 }$ &	${\bf 0.418\pm0.022}$&	 ${\bf 0.210 \pm 0.004}$  & ${\bf 2.862\pm0.166}$ & ${\bf 0.245\pm0.019}$\\ 
 \bottomrule
 \end{tabular}
\end{center}
\label{table:1}
\vspace{-.3in}
\end{table} 

Table \ref{table:1} compares the predictive performance of the mortality and length of stay prediction task on MIMIC-III. The proposed model consistently achieves the best average score over all the metrics. 
We note that a paired t-test indicates that the proposed model results in statistically significant improvements over all baseline models $(p<0.01)$ with respect to all the metrics except median absolute error. 
The version of the proposed model used in this experiment includes all three interpolation network
outputs (smooth interpolation, transients, and intensity function).

\section{Discussion and Conclusions}

In this paper, we have presented a new framework for dealing with the
problem of supervised learning in the presence of sparse
and irregularly sampled time series. The proposed framework is fully
modular. It uses an interpolation network to accommodate the complexity
that results from using sparse and irregularly sampled data as
supervised learning inputs, followed by the application of 
a prediction network that operates over the regularly spaced and fully
observed, multi-channel output provided by the interpolation network.
The proposed approach also addresses some difficulties with prior 
approaches including the complexity of the Gaussian process
interpolation layers used in \citet{li2016scalable}, and 
the lack of modularity in the approach of \citet{che2016recurrent}.
Our framework also introduces novel elements including the
use of semi-parametric, feed-forward interpolation layers,
and the decomposition of an irregularly sampled 
input time series into multiple distinct information channels.
Our results show statistically significant improvements for both classification and regression tasks
 over a range of
baseline and state-of-the-art methods. 

\subsection*{Acknowledgements}

This work was supported by the National Science Foundation under 
Grant No.~1350522.

\bibliographystyle{plain}
\bibliography{iclr2019_conference}

\begin{thebibliography}{15}
\providecommand{\natexlab}[1]{#1}
\providecommand{\url}[1]{\texttt{#1}}
\expandafter\ifx\csname urlstyle\endcsname\relax
  \providecommand{\doi}[1]{doi: #1}\else
  \providecommand{\doi}{doi: \begingroup \urlstyle{rm}\Url}\fi

\bibitem[Breiman(2001)]{breiman_rf}
Leo Breiman.
\newblock Random forests.
\newblock \emph{Mach. Learn.}, 45\penalty0 (1):\penalty0 5--32, October 2001.
\newblock ISSN 0885-6125.
\newblock \doi{10.1023/A:1010933404324}.
\newblock URL \url{https://doi.org/10.1023/A:1010933404324}.

\bibitem[Che et~al.(2018)Che, Purushotham, Cho, Sontag, and
  Liu]{che2016recurrent}
Zhengping Che, Sanjay Purushotham, Kyunghyun Cho, David Sontag, and Yan Liu.
\newblock Recurrent neural networks for multivariate time series with missing
  values.
\newblock \emph{Scientific Reports}, 8\penalty0 (1):\penalty0 6085, 2018.
\newblock URL \url{https://doi.org/10.1038/s41598-018-24271-9}.

\bibitem[Cortes and Vapnik(1995)]{cortes1995}
Corinna Cortes and Vladimir Vapnik.
\newblock Support-vector networks.
\newblock \emph{Machine Learning}, 20\penalty0 (3):\penalty0 273--297, Sep
  1995.
\newblock ISSN 1573-0565.
\newblock \doi{10.1007/BF00994018}.
\newblock URL \url{https://doi.org/10.1007/BF00994018}.

\bibitem[Davis and Goadrich(2006)]{auprc}
Jesse Davis and Mark Goadrich.
\newblock The relationship between precision-recall and roc curves.
\newblock In \emph{Proceedings of the 23rd International Conference on Machine
  Learning}, ICML '06, pages 233--240, New York, NY, USA, 2006. ACM.
\newblock ISBN 1-59593-383-2.
\newblock \doi{10.1145/1143844.1143874}.
\newblock URL \url{http://doi.acm.org/10.1145/1143844.1143874}.

\bibitem[Freund and Schapire(1997)]{adaboost}
Yoav Freund and Robert~E Schapire.
\newblock A decision-theoretic generalization of on-line learning and an
  application to boosting.
\newblock \emph{J. Comput. Syst. Sci.}, 55\penalty0 (1):\penalty0 119--139,
  August 1997.
\newblock ISSN 0022-0000.
\newblock \doi{10.1006/jcss.1997.1504}.
\newblock URL \url{http://dx.doi.org/10.1006/jcss.1997.1504}.

\bibitem[Futoma et~al.(2017)Futoma, Hariharan, Heller, Sendak, Brajer, Clement,
  Bedoya, and O’Brien]{futoma2017improved}
Joseph Futoma, Sanjay Hariharan, Katherine Heller, Mark Sendak, Nathan Brajer,
  Meredith Clement, Armando Bedoya, and Cara O’Brien.
\newblock An improved multi-output gaussian process rnn with real-time
  validation for early sepsis detection.
\newblock In \emph{Machine Learning for Healthcare Conference}, pages 243--254,
  2017.

\bibitem[Hastie et~al.(2001)Hastie, Tibshirani, and
  Friedman]{hastie01statisticallearning}
Trevor Hastie, Robert Tibshirani, and Jerome Friedman.
\newblock \emph{The Elements of Statistical Learning}.
\newblock Springer Series in Statistics. Springer New York Inc., New York, NY,
  USA, 2001.

\bibitem[Hosmer~Jr et~al.(2013)Hosmer~Jr, Lemeshow, and
  Sturdivant]{hosmer2013applied}
David~W Hosmer~Jr, Stanley Lemeshow, and Rodney~X Sturdivant.
\newblock \emph{Applied logistic regression}, volume 398.
\newblock John Wiley \& Sons, 2013.

\bibitem[Johnson et~al.(2016)Johnson, Pollard, Shen, Li-wei, Feng, Ghassemi,
  Moody, Szolovits, Celi, and Mark]{johnson2016mimic}
Alistair~EW Johnson, Tom~J Pollard, Lu~Shen, H~Lehman Li-wei, Mengling Feng,
  Mohammad Ghassemi, Benjamin Moody, Peter Szolovits, Leo~Anthony Celi, and
  Roger~G Mark.
\newblock Mimic-iii, a freely accessible critical care database.
\newblock \emph{Scientific data}, 3:\penalty0 160035, 2016.

\bibitem[Lasko(2014)]{lasko2014efficient}
Thomas~A Lasko.
\newblock Efficient inference of gaussian-process-modulated renewal processes
  with application to medical event data.
\newblock In \emph{Uncertainty in artificial intelligence: proceedings of
  the... conference. Conference on Uncertainty in Artificial Intelligence},
  volume 2014, page 469. NIH Public Access, 2014.

\bibitem[Li and Marlin(2016)]{li2016scalable}
Steven Cheng-Xian Li and Benjamin~M Marlin.
\newblock A scalable end-to-end gaussian process adapter for irregularly
  sampled time series classification.
\newblock In \emph{Advances In Neural Information Processing Systems}, pages
  1804--1812, 2016.

\bibitem[Li and Marlin(2015)]{li2015classification}
Steven Cheng-Xian Li and Benjmain~M. Marlin.
\newblock Classification of sparse and irregularly sampled time series with
  mixtures of expected {G}aussian kernels and random features.
\newblock In \emph{31st Conference on Uncertainty in Artificial Intelligence},
  2015.

\bibitem[Lipton et~al.(2016)Lipton, Kale, and Wetzel]{lipton2016directly}
Zachary~C Lipton, David Kale, and Randall Wetzel.
\newblock Directly modeling missing data in sequences with rnns: Improved
  classification of clinical time series.
\newblock In \emph{Machine Learning for Healthcare Conference}, pages 253--270,
  2016.

\bibitem[Marlin et~al.(2012)Marlin, Kale, Khemani, and Wetzel]{marlin-ihi2012}
Benjamin~M. Marlin, David~C. Kale, Robinder~G. Khemani, and Randall~C. Wetzel.
\newblock Unsupervised pattern discovery in electronic health care data using
  probabilistic clustering models.
\newblock In \emph{Proceedings of the 2nd ACM SIGHIT International Health
  Informatics Symposium}, pages 389--398, 2012.

\bibitem[Yadav et~al.(2018)Yadav, Steinbach, Kumar, and Simon]{yadav2018mining}
Pranjul Yadav, Michael Steinbach, Vipin Kumar, and Gyorgy Simon.
\newblock Mining electronic health records (ehrs): A survey.
\newblock \emph{ACM Computing Surveys (CSUR)}, 50\penalty0 (6):\penalty0 85,
  2018.

\end{thebibliography}

\appendix
\section{Appendix}
\subsection{Model Architecture}

\label{diagram}
\begin{figure}[H]
\includegraphics[width=\linewidth]{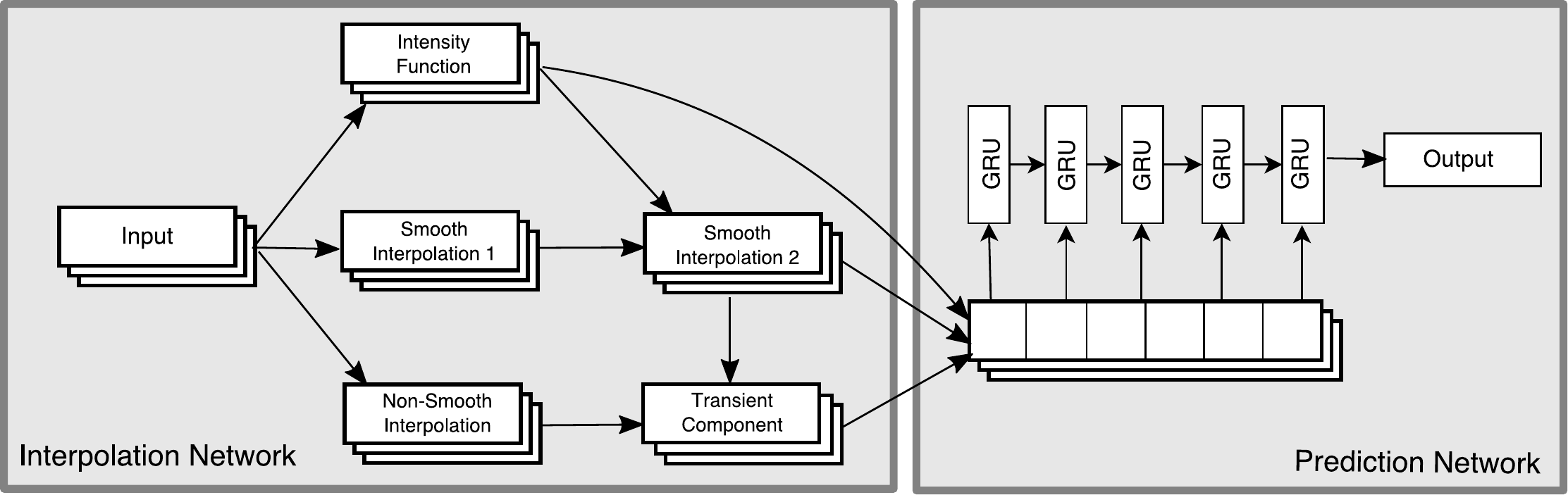}
\centering
\caption{Architecture of the proposed model}
\label{fig:model}
\end{figure}

\subsection{Dataset Description}
\label{dataset}

We evaluate our model framework on the publicly available MIMIC-III dataset \citep{johnson2016mimic}. MIMIC-III is a de-identified
dataset collected at Beth Israel Deaconess Medical Center
from 2001 to 2012. It consists of approximately 58,000 hospital admission records.  This data set contains sparse and irregularly sampled physiological signals, medications, diagnostic codes, in-hospital mortality, length of stay and more. We focus on predicting in-hospital mortality and length of stay using the first 48 hours of data. We extracted 12 standard physiological variables from each of the 53,211 records obtained after removing hospital admission records with length of stay less than 48 hours. Table \ref{table:mis} shows the features, sampling rates (per hour) and their missingness information computed using the union of all time stamps that exist in any dimension of the input time series.

\begin{table}[h]
\centering
\caption{Features extracted from MIMIC III for our experiments}
\label{table:mis}
                \begin{tabular}[h]{l c c}
                
                 \toprule
                 {feature} & {\#Missing} & {Sampling Rate}\\
                    \midrule
                    SpO2 & $31.35\%$ & $0.80$\\
                    HR  & $23.23\%$ & $0.90$\\
                    RR & $59.48\%$  & $0.48$\\
                    SBP & $49.76\%$ & $0.59$\\
                    DBP & $48.73\%$ & $0.60$\\
                    Temp & $83.80\%$& $0.19$\\
  
                    \bottomrule
                    \end{tabular}  
                    \,\,\,\,\,\,
                \begin{tabular}[h]{l c c}
                 \toprule
                 {feature} & {\#Missing} & {Sampling Rate}\\
                    \midrule
                    TGCS & $87.94\%$& $0.14$\\
                    CRR &$95.08\%$& $0.06$\\
                    UO & $82.47\%$&$0.20$\\
                    FiO2 &$94.82\%$&$0.06$\\
                    Glucose &$91.47\%$&$0.10$\\
                    pH &$96.25\%$&$0.04$\\
                    \bottomrule
                    \end{tabular}
                \end{table}

\cut{
\begin{table}[h]
\caption{Features extracted from MIMIC III for our experiments}
\label{table:mis}
            \footnotesize
                \begin{tabular}[h]{l c}
                 \toprule
                 {feature} & {\#Missing}\\
                    \midrule
                    SpO2 & $31.35\%$\\
                    HR  & $23.23\%$\\
                    RR & $59.48\%$\\
                    
                    \bottomrule
                \end{tabular}
               \hfill
                \begin{tabular}[h]{l c}
                 \toprule
                 {feature} & {\#Missing}\\
                    \midrule
                    SBP & $49.76\%$\\
                    DBP & $48.73\%$\\
                    Temp & $83.80\%$\\
  
                    \bottomrule
                    \end{tabular}
                 \hfill
                \begin{tabular}[h]{l c}
                 \toprule
                 {feature} & {\#Missing}\\
                    \midrule
                    TGCS & $87.94\%$\\
                    CRR &$95.08\%$\\
                    UO & $82.47\%$\\
                    
                    \bottomrule
                    \end{tabular}
                     \hfill
                \begin{tabular}[h]{l c}
                 \toprule
                {feature} & {\#Missing}\\
                    \midrule
                    FiO2 &$94.82\%$\\
                    Glucose &$91.47\%$\\
                    pH &$96.25\%$\\
                    \bottomrule
                    \end{tabular}
                \end{table}
}
\subsubsection*{Prediction Tasks}

In our experiments, each admission record corresponds to one data case $(\mbf{s_n},y_n)$. Each data case $n$ consists of a sparse and irregularly sampled time series $\mbf{s}_n$ with $D=12$ dimensions. Each dimension $d$ 
of $\mbf{s}_n$ corresponds to one of the 12 vital sign time series mentioned above. In the case of classification, $y_n$ is a binary indicator where $y_n=1$ indicates that the patient died at any point within the hospital stay following the first 48 hours and $y_n=0$ indicates that the patient was discharged at any point after the first 48 hours. There are 4310 (8.1\%) patients with a $y_n=1$ mortality label. The complete data set is $\mathcal{D}=\{(\mbf{s_n},y_n)|n=1,...,N\}$, and there
are $N=53,211$ data cases. The goal in the classification task is to learn a classification function $h$ of the
form $\hat{y}_n \leftarrow h(\mbf{s}_n)$ where $\hat{y}_n$ is a discrete value.

In the case of regression, $y_n$ is a real-valued regression target corresponding to the length of stay. Since the
data set include some very long stay durations, we let $y_n$ represent the log of the length of stay in days for 
all models. We convert back from the log number of days to the number of days when reporting results.
The complete data set is again $\mathcal{D}=\{(\mbf{s_n},y_n)|n=1,...,N\}$ with
$N=53,211$ data cases (we again require 48 hours worth of data). The goal in the regression task is to learn a regression function $h$ of the form $\hat{y}_n \leftarrow h(\mbf{s}_n)$ where $\hat{y}_n$ is a continuous value.

\subsection{Implementation Details}
\label{implementation}
\subsubsection{Proposed Model}
The model is learned using the Adam optimization method 
in TensorFlow with gradients provided via automatic differentiation.
However, the actual multivariate time series  
representation used during learning is based on the union
of all time stamps that exist in any dimension of the input time series.
Undefined observations are represented as zeros and a separate
missing data mask is used to keep track of which time series have
observations at each time point. Equations 1 \& 2 are modified such
that data that are not available are not taken into account at all.
This implementation is exactly equivalent to the computations
described in Equations 1 \& 2, but support parallel computation
across all dimensions of the time series for a given data case.

Finally, we note that the learning problem can be solved using a
doubly stochastic gradient based on the use of mini batches
combined with re-sampling the artificial missing data masks 
used in the interpolation loss. In practice, we randomly select
$20\%$ of the observed data points to hold out from every input time
series.

\subsubsection{Baselines}
The Logistic Regression model is trained with cross entropy loss with regularization strength set to 1. The support vector classifier is used with a RBF kernel and trained to minimize the soft margin loss. We use the cross entropy loss on the validation set to select the optimal number of estimators in case of Adaboost and Random Forest. Similar to the classification setting, the optimal number of estimators for regression task in Adaboost and Random Forest is chosen on the basis of squared error on validation set.

We evaluate all models using a five-fold cross-validation estimate of generalization performance.
In the classification setting,  all the deep learning baselines are trained to minimize the cross entropy loss while the proposed model uses a composite loss consisting of cross-entropy loss and interpolation loss (with $\delta = 1$) as described in section $3.2.3$. In the case of the regression task, all baseline models are trained to minimize squared error and the proposed model is again trained with a composite loss consisting of squared error and interpolation loss.

For all of the GRU-based models, we use the already specified parameters \citep{che2016recurrent}. The models are learned using the Adam optimization. Early
stopping is used on a validation set sub-sampled from the training folds. In the classification case,
the final outputs of the GRU hidden units are used in a logistic layer that predicts the class.
In the regression case, the final outputs of the GRU hidden units
are used as input for a dense hidden layer with $50$ units, followed by a linear output layer.

\end{document}